\def\year{2020}\relax
\documentclass[letterpaper]{article}
\usepackage{aaai20}
\usepackage{times}
\usepackage{helvet}
\usepackage{courier}
\usepackage{url}
\usepackage{graphicx}
\usepackage{microtype}
\usepackage{subfig}
\usepackage{booktabs}
\usepackage{mathtools}
\mathtoolsset{showonlyrefs,showmanualtags}
\usepackage{bm}
\usepackage{bigints}
\usepackage{amsfonts}
\usepackage{amsthm}
\usepackage{dsfont}
\usepackage{empheq}
\usepackage{xparse}
\usepackage{multirow}
\usepackage{array}
\usepackage{mathrsfs}
\allowdisplaybreaks

\newcommand{\bu}{\mathbf{u}}

\newcommand{\KL}{\text{KL}}
\newcommand{\CELBO}{\text{CELBO}}
\newcommand{\intd}{\text{ d}}
\newcommand{\bS}{\mathbf{S}}
\newcommand{\bK}{\mathbf{K}}

\newcommand{\nn}{\nonumber}

\newcommand{\mT}{\mathcal T}
\newcommand{\mD}{\mathcal D}
\newcommand{\citet}[1]{\citeauthor{#1} \shortcite{#1}} \newcommand{\citep}{\cite} 

\DeclareMathOperator{\E}{\mathbb{E}}
\DeclareMathOperator{\Var}{\text{Var}}

\DeclareMathOperator{\Tr}{\text{Tr}}

\newtheorem{theorem}{Theorem}
\theoremstyle{remark}
\newtheorem*{remark}{Remark}

\title{Variational Inference for Sparse Gaussian Process Modulated Hawkes Process}
\author{
Rui Zhang,\textsuperscript{\rm 1,2}
Christian Walder,\textsuperscript{\rm 1,2}
Marian-Andrei Rizoiu\textsuperscript{\rm 3}\\
\textsuperscript{\rm 1}The Australian National University,
\textsuperscript{\rm 2}Data61 CSIRO,
\textsuperscript{\rm 3}University of Technology Sydney\\
Rui.Zhang@anu.edu.au, Christian.Walder@data61.csiro.au, Marian-Andrei.Rizoiu@uts.edu.au
}

\begin{document}

\maketitle

\begin{abstract}
The Hawkes process (HP) has been widely applied to modeling self-exciting events including neuron spikes, earthquakes and tweets.
To avoid designing parametric triggering kernel and to be able to quantify the prediction confidence, the non-parametric Bayesian HP has been proposed.
However, the inference of such models suffers from unscalability or slow convergence.
In this paper, we aim to solve both problems. Specifically, first, we propose a new non-parametric Bayesian HP in which the triggering kernel is modeled as a squared sparse Gaussian process.
Then, we propose a novel variational inference schema for model optimization. We employ the branching structure of the HP so that maximization of evidence lower bound (ELBO) is tractable by the expectation-maximization algorithm. We propose a tighter ELBO which improves the fitting performance. 
Further, we accelerate the novel variational inference schema to linear time complexity by leveraging the stationarity of the triggering kernel. Different from prior acceleration methods, ours enjoys higher efficiency. Finally, we exploit synthetic data and two large social media datasets to evaluate our method. We show that our approach outperforms state-of-the-art non-parametric frequentist and Bayesian methods. We validate the efficiency of our accelerated variational inference schema and practical utility of our tighter ELBO for model selection. We observe that the tighter ELBO exceeds the common one in model selection. 
\end{abstract}

\section{Introduction}
\label{sec:introduction}
\begin{figure}[t!]
    \subfloat[Poisson Cluster Process ]{\includegraphics[width=0.58\columnwidth]{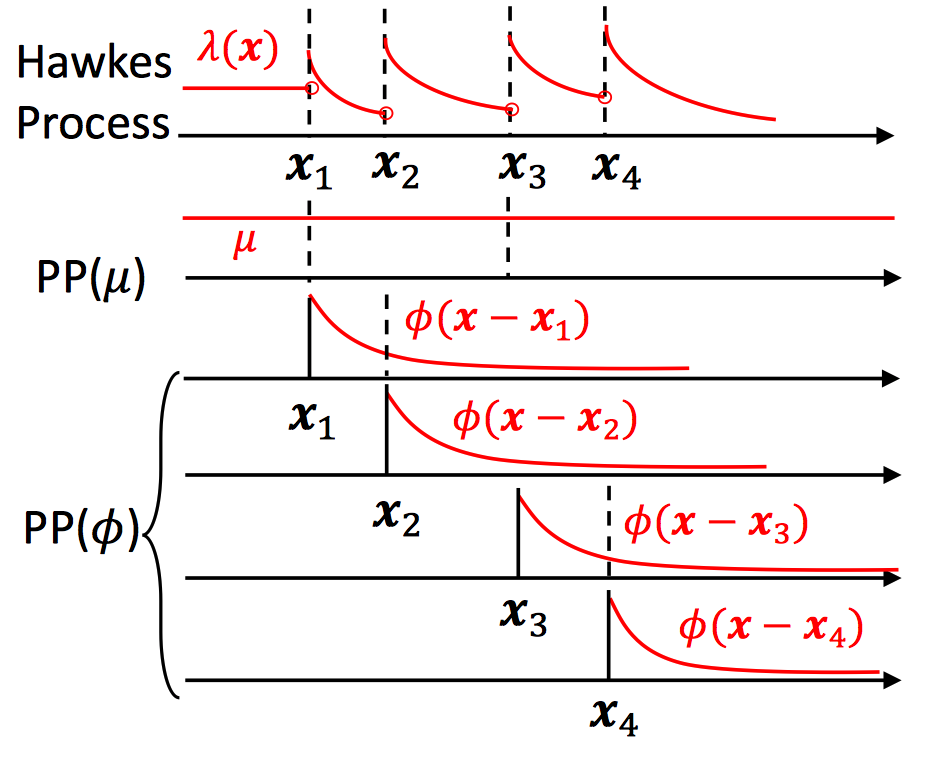}\label{fig:poisson-cluster}}
    \subfloat[Branching Structure]{\includegraphics[width=0.4\columnwidth]{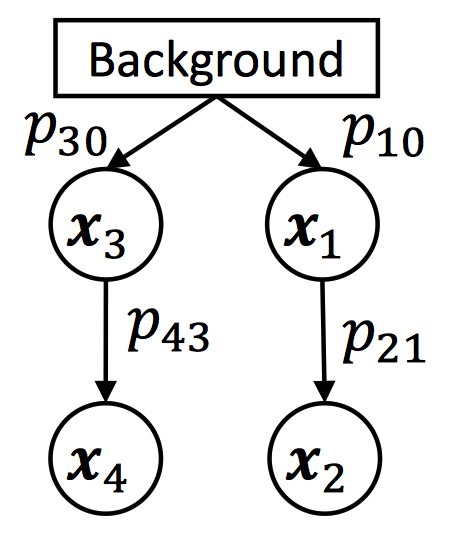}\label{fig:branching-structure}}
    \caption{
        The Cluster Representation of a HP.
        (a) A HP with a decaying triggering kernel $\phi(\cdot)$ has intensity $\lambda(\bm x)$ which increases after each new point (dash line) is generated. It can be represented as a cluster of PPes: PP($\mu$) and PP($\phi(\bm x-\bm x_i)$) associated with each $\bm x_i$.
        (b) The branching structure corresponding to the triggering relationships shown in (a),
        where an edge $\bm x_i \rightarrow \bm x_j$ means that $\bm x_i$ triggers $\bm x_j$, and its probability is denoted as $p_{ji}$.
    }
    \label{fig:hp_2_pps}
\end{figure}

The Hawkes process (HP) \citep{hawkes1971} is particularly useful to model self-exciting point data -- i.e., when the occurrence of a point increases the likelihood of occurrence of new points.
The process is parameterized using a background intensity $\mu$, and a triggering kernel $\phi$.
The Hawkes process can be alternatively represented as a cluster of Poisson processes (PPes) \citep{hawkes1974cluster}. In the cluster, a PP with an intensity $\mu$ (denoted as PP($\mu$)) generates immigrant points which are considered to arrive in the system from the outside, and every existing point triggers offspring points, which are generated internally through the self-excitement, following a PP($\phi$).
Points can therefore be structured into clusters where each cluster contains either a point and its direct offspring or the background process (an example is shown in Fig. \ref{fig:poisson-cluster}). Connecting all points using the triggering relations yields a tree structure, which is called the branching structure (an example is shown in Fig. \ref{fig:branching-structure} corresponding to Fig.\ref{fig:poisson-cluster}). With the branching structure, we can decompose the HP into a cluster of PPes.
The triggering kernel $\phi$ is shared among all cluster Poisson processes relating to a HP, and it determines the overall behavior of the process.
Consequently, designing the kernel functions is of utmost importance for employing the HP to a new application, and its study
has attracted much attention.

\textbf{Prior non-parametric frequentist solutions.}
In the case in which the optimal triggering kernel for a particular application is unknown, a typical solution is to express it using a
non-parametric form, such as the work of \citet{lewis2011nonparametric}; \citet{zhou2013learning}; \citet{bacry2014second}; \citet{eichler2017graphical}.
These are all frequentist methods and among them, the Wiener-Hoef equation based method \citep{bacry2014second} enjoys linear time complexity. Our method has a similar advantage of linear time complexity per iteration.  The Euler-Lagrange equation based solutions \citep{lewis2011nonparametric,zhou2013learning} require discretizing the input domain so they face a problem of poorly scaling with the dimension of the domain. The same problem is also faced by \citet{eichler2017graphical}'s discretization based method.
In contrast, our method requires no dixscretization so enjoys scalability with the dimension of the domain.

\textbf{Prior Bayesian solutions.} The Bayesian inference for the HP has also been studied, including the work of \citet{rasmussen2013bayesian}; \citet{linderman2014discovering}; \citet{linderman2015scalable}. These work require either constructing a parametric triggering kernel \citep{rasmussen2013bayesian,linderman2014discovering}  or discretizing the input domain to scale with the data size \citep{linderman2015scalable}.
The shortcoming of discretization is just mentioned and to overcome it, \citet{donnet2018nonparametric} propose a continuous non-parametric Bayesian HP and resort to an unscalable Markov chain Monte Carlo (MCMC) estimator to the posterior distribution. 
A more recent solution based on Gibbs sampling~\citep{zhang2018efficient} obtains linear time complexity per iteration by exploiting the HP's branching structure and the stationarity of the triggering kernel similar to the work of \citet{halpin2012algorithm}. This approach considers only high-probability triggering relationships in computations for acceleration. However, those relationships are updated in each iteration, which is less efficient than our pre-computing them. Besides, our variational inference schema enjoys faster convergence than that of the Gibbs sampling based method.

\textbf{Contributions.}
In this paper, we propose the first sparse Gaussian process modulated HP which employs a novel variational inference schema, enjoys a linear time complexity per iteration and scales to large real world data. Our method is inspired by the variational Bayesian PP (VBPP) \citep{lloyd2015variational} which provides the Bayesian non-parametric inference only for the whole intensity of the HP without for its components: the background intensity $\mu$ and the triggering kernel $\phi$. Thus, the VBPP loses the internal reactions between points, and developing the variational Bayesian non-parametric inference for the HP is non-trivial and more challenging than the VBPP. In this paper, we adapt the VBPP for the HP and term the new approach the variational Bayesian Hawkes process (VBHP). The contributions are summarized:

\textbf{(1)} (Sec.\ref{sec:vbhp}) We introduce a new Bayesian non-parametric HP which employs a sparse Gaussian process modulated triggering kernel and a Gamma distributed background intensity.

\textbf{(2)} (Sec.\ref{sec:vbhp}\&\ref{sec:new_vi}) We propose a new variational inference schema for such a model. Specifically, we employ the branching structure of the HP so that maximization of the evidence lower bound (ELBO) is tractable by the expectation-maximization algorithm, and we contribute a tighter ELBO which improves the fitting performance of our model. 

\textbf{(3)} (Sec.\ref{sec:acceleration}) We propose a new acceleration trick based on the stationarity of the triggering kernel. The new trick enjoys higher efficiency than prior methods and accelerates the variational inference schema to linear time complexity per iteration. 

\textbf{(4)} (Sec.\ref{sec:experiments}) We empirically show that VBHP provides more accurate predictions than state-of-the-art methods on synthetic data and on two large online diffusion datasets. We validate the linear time complexity and faster convergence of our accelerated variational inference schema compared to the Gibbs sampling method, and the practical utility of our tighter ELBO for model selection, which outperforms the common one in model selection.

\section{Prerequisite}
In this section, we review the Hawkes process, the variational inference and its application to the Poisson process.

\subsection{Hawkes Process (HP)}
\label{sec:hawkes_process}

The HP \citep{hawkes1971} is a self-exciting point process, in which the occurrence of a point increases the arrival rate $\lambda(\cdot)$, a.k.a. the 
(conditional) intensity, of new points. 
Given a set of time-ordered points $\mD = \{\bm x_i\}_{i=1}^{N}$, $\bm x_i \in \mathbb{R}^{R}$, the  intensity at $\bm x$ conditioned on given points is written as:
\begin{equation}
    \lambda(\bm x) = \mu + \sum_{\bm x_i < \bm x}\phi(\bm x-\bm x_i),
    \label{eq:intensity}
\end{equation}
where $\mu>0$ is a constant background intensity, and $\phi: \mathbb{R}^{R} \rightarrow [0, \infty)$ is the triggering kernel.
 
We are particularly interested in the branching structure of the HP. As introduced in Sec.\ref{sec:introduction}, each point $\bm x_i$ has a parent that we represent by a one-hot vector $\bm b_i=[ b_{i0}, b_{i1},\cdots, b_{i,i-1}]^T$: each element $b_{ij}$ is binary, $ b_{ij}=1$ represents that $\bm x_i$ is triggered by $\bm x_j$ ($0 \leq j\leq i-1$, $\bm x_0$: the background), and $\sum_{j=0}^{i-1}  b_{ij}=1$. A branching structure $B$ is a set of $\bm b_i$, namely $B=\{\bm b_i\}_{i=1}^{N}$, and if the probability of $b_{ij}=1$ is defined as $p_{ij} \equiv p(b_{ij}=1)$, the probability of $B$ can be expressed as $p(B) = \prod_{i=1}^{N} \prod_{j=0}^{i-1} p_{ij}^{b_{ij}}$. Since $\bm b_i$ is a one-hot vector, there is $\sum_{j=0}^{i-1}p_{ij}=1$ for all $i$.

Given both $\mD$ and a branching structure $B$, the log likelihood of $\mu$ and $\phi$ 
becomes a sum of log likelihoods of PPes. With $B$, the HP $\mD$ can be decomposed into PP($\mu$) and $\{\text{PP}(\phi(\bm x - \bm x_i))\}_{i=1}^{N}$. The data domain of PP($\mu$) equals that of the HP, which we denote $\mT$,
and we denote the data domain of PP($\phi(\bm x - \bm x_i)$) by $\mT_i \subset \mT$. As a result, the log likelihood $\log p(D,B|\mu,\phi)$ is expressed as:
{\medmuskip=1mu
	\thinmuskip=1mu
	\thickmuskip=1mu
\begin{align}
&
\log p(\mD, B | \mu,\phi) =
\\ &
\sum_{i=1}^{N}\Big(\sum_{j=1}^{i-1} b_{ij} \log \phi_{ij}+b_{i0}\log \mu \Big)
 -\sum_{i=1}^{N}\int_{\mT_i}\phi -\mu |\mT|,~~\label{eq:log_db}
\end{align}}
where $|\mT| \equiv \int_{\mT}1 \intd \bm x$,  $\phi_{ij}\equiv \phi(\bm x_i - \bm x_j)$ and $\int_{\mT_i}\phi \equiv \int_{\mT_i}\phi(\bm x) \intd \bm x$. Throughout this paper, we simplify the integration by omitting the integration variable, which is $\bm x$ unless otherwise specified.
This work is developed based on the univariate HP. However, it can be extended to be multivariate following the same procedure.

\subsection{Variational Inference (VI)}\label{sec:vi}
Consider the latent variable model $p(\bm x, \bm z| \bm \theta)$ where $\bm x$ and $\bm z$ are the data and the latent variables respectively. The variational approach introduces a variational distribution to approximate the posterior distribution $q(\bm z  |\bm \theta') \approx p(\bm z | \bm x, \bm \theta)$ and maximizes a lower bound of the log-likelihood, which can be derived from the non-negative gap perspective:
\begin{align}
&\log p(\bm x | \bm \theta)\\
&=\log \dfrac{p(\bm x, \bm z | \bm \theta)}{q(\bm z |\bm \theta')}  - \log \dfrac{p(\bm z |\bm x, \bm \theta)}{q(\bm z | \bm \theta')}  \\
&=\underbrace{\underbrace{\E_{q(\bm z |\bm \theta' )} \big [\log p(\bm x | \bm z, \bm \theta) \big ]}_{\text{reconstruction term}} - \underbrace{\KL (q(\bm z| \bm \theta')  || p(\bm z | \bm \theta) )}_{\text{regularisation term}} }_{\equiv \CELBO(q(\bm z),p(\bm x| \bm z), p(\bm z ))} \\
&\quad +\underbrace{\KL  (q(\bm z|\bm \theta') | |p (\bm z | \bm x, \bm \theta) )}_{\substack{ \text{intractable (non-negative) gap}}}  \\
&\geq \CELBO(q(\bm z),p(\bm x| \bm z), p(\bm z )), \label{eq:vi_elbo}
\end{align}
where we omit $\bm \theta$ and $\bm \theta'$ in conditions. We term this the Common ELBO (CELBO) to differentiate with our tighter ELBO of Sec.\ref{sec:new_vi}. For notational convenience, we will often omit conditioning on $\bm \theta$ and $\bm \theta'$ hereinafter.
Optimizing the CELBO w.r.t. $\bm \theta'$ balances between the reconstruction error and the Kullback-Leibler (KL) divergence from the prior.  Generally, the conditional $p(\bm x | \bm z)$ is known, so is the prior. Thus, for an appropriate choice of $q$, it is easier to work with this lower bound than with  the intractable posterior $p(\bm z | \bm x)$. We also see that, due to the form of the intractable gap, if $q$ is from a distribution family containing elements close to the true unknown posterior, then $q$ will be close to the true posterior when the CELBO is close to the true likelihood.
An alternative derivation applies Jensen's inequality \citep{jordan1999introduction}.

\subsection{Variational Bayesian Poisson Process (VBPP)}
\label{sec:vbpp}
VBPP \citep{lloyd2015variational} applies the VI to the Bayesian Poisson process, which exploits the sparse Gaussian process (GP) to model the Poisson intensity. Specifically, VBPP uses a squared link function to map a sparse GP distributed function $f$ to the Poisson intensity $\lambda(\bm x) = f^2(\bm x)$. The sparse GP employs the ARD kernel:
\begin{equation}
    K(\bm x,\bm x') \equiv \gamma \prod_{r=1}^{R} \exp\Big( -\dfrac{(x_r - x_r')^2}{2 \alpha_r} \Big ).
\end{equation} 
where $\gamma$ and $\{\alpha_r\}_{r=1}^{R}$ are GP hyper-parameters. Let $\bm u \equiv (f(\bm z_1),f(\bm z_2),\cdots,f(\bm z_M))$ where $\bm z_i$ are inducing points. The prior and the approximate posterior distributions of $\bm u$ are Gaussian distributions $p(\bm u ) = \mathcal N(\bm u | \bm 0,\bm K_{zz})$ and $q(\bm u ) = \mathcal N(\bm u | \bm  m,\bm S)$ where $\bm m$ and $\bS$ are the mean vector and the covariance matrix respectively. Note both $\bm u$ and $f$ employ zero mean priors. Notations of VBPP are connected with those of VI (Sec.\ref{sec:vi}) in Table\ref{tab:notations}.
\begin{table}
\centering
\caption{Notations}
\label{tab:notations}
	\begin{tabular}{ccc}
		\toprule
		VBHP & VBPP & VI \\
		\midrule
		$\mD \equiv \{\bm x_n\}_{n=1}^{N}$ & $\mD \equiv \{\bm x_n\}_{n=1}^{N}$
		 & $\bm x$ 
		 \\
		$B, \mu, f, \bm u$& $f, \bm u$ & $\bm z$ \\
		$k_0,c_0,\{\alpha_i\}_{i=1}^{R},\gamma$
		& $\{\alpha_i\}_{i=1}^{R},\gamma$  
		& $\bm \theta$ 
		\\
		$k,c,\bm m, \bm S ,\{\alpha_i\}_{i=1}^{R},\gamma,$
		& \multirow{2}{*}{$\bm m, \bm S,\{\alpha_i\}_{i=1}^{R},\gamma$}
		& \multirow{2}{*}{$\bm \theta'$} 
		\\
		$\{\{q_{ij}\}_{j=0}^{i-1}\}_{i=1}^{N}$
		& 
		&  \\
		\bottomrule
	\end{tabular}
\end{table}

Importantly, the variational joint distribution of $f$ and $\bm u$ uses the exact conditional distribution $p(f|\bm u)$, i.e., 
\begin{equation}\label{eq:exact_conditional_vbpp}
    q(f, \bm u) \equiv p(f | \bm u)q(\bm u)
\end{equation}
which in turn leads to the posterior GP:
\begin{align} \label{eq:post_f}
    q(f)&= \mathcal N( f | \nu,\Sigma), \\
    \nu(\bm x) &\equiv \bm K_{xz} \bm K_{zz}^{-1}\bm m,\\
    \Sigma(\bm x, \bm x')&\equiv\bm K_{xx'}+\bm K_{xz} \bm K_{zz}^{-1}(\bm S \bm K_{zz}^{-1}-I)\bm K_{zx'}.
\end{align}
Then, the CELBO is obtained by using Eqn.\eqref{eq:vi_elbo}:
\begin{align}
    &\CELBO(q(f, \bm u),p(\mD | f, \bm u),p(f, \bm u )) \\
    &= \E_{q( f)}  [ \log p(\mD|f) ]- \KL  ( q(\bm u) || p(\bm u)  ). \label{eq:vbpp_elbo}
\end{align}
Note that the second term is the KL divergence between two multivariate Gaussian distributions, so is available in closed form. The first term turns out to be the expectation w.r.t. $q(f)$ of the log-likelihood $\log p(\mD | f) = \sum_{i=1}^{N}\log f^2(\bm x_i) - \int_{\mT} f^2$. The expectation of the integral part is relatively straight-forward to compute and the expectation of the other (data-dependent) part is available in almost closed-form with a hyper-geometric function.


\section{Variational Bayesian Hawkes Process}
\label{sec:vbhp}
\subsection{Notations} 
To extend VBPP to HP, we introduce two more variables: the background intensity $\mu$ and the branching structure $B$, defined in Sec.\ref{sec:hawkes_process}. We assume that the prior distribution of $\mu$ is a Gamma distribution $p(\mu)=\text{Gamma}(\mu|k_0,c_0)$\footnote{$\text{Gamma}(\mu|k_0,c_0)=\dfrac{1}{\Gamma(k_0)c_0^k}\mu^{k_0-1}e^{-x/c_0}$} and the posterior distribution is approximated by another Gamma distribution $q(\mu)= \text{Gamma}(\mu | k,c)$. For $B=\{\bm b_i\}_{i=1}^N$ given $\mD=\{\bm x_i\}_{i=1}^N$, we assume the variational posterior $b_{ij}$, $j=0,\cdots, i-1$, have a categorical distribution: $q_{ij} = q(b_{ij}=1)$ and $\sum_{j=0}^{i-1}q_{ij}=1$, and thus, the variational posterior probability of $B$ is expressed as:
\begin{align}
    q(B) = \prod_{i=1}^N \prod_{j=0}^{i-1}q_{ij}^{b_{ij}}.\label{eq:q_b}
\end{align}

The same squared link function is adopted for the triggering kernel $\phi(\bm x) = f^2(\bm x)$, so are the priors for $f$ and $\bm u$, namely $\mathcal N(f|\bm 0, \bK_{xx'})$ and $\mathcal N(\bm u|\bm 0, \bK_{zz'})$. More link functions such as $\exp(\cdot)$ are discussed by \citet{lloyd2015variational}. Moreover, we use the same variational joint posterior on $f$ and $\bm u$ as Eqn.\eqref{eq:exact_conditional_vbpp}.
Consequently, we complete the variational joint distribution on all latent variables as below:
{\medmuskip=1mu
	\thinmuskip=1mu
	\thickmuskip=1mu
\begin{equation}\label{eq:vbhp_approx_joint_posterior}
	q(B, \mu, f, \bm u)\equiv q(B) q(\mu) p(f| \bm u)q(\bm u),
\end{equation}}
and notations of VBHP are summarized in Table \ref{tab:notations}.

\subsection{CELBO} 
Based on Eqn.\eqref{eq:vi_elbo}\&\eqref{eq:vbhp_approx_joint_posterior}, we obtain the CELBO for VBHP (see details in A.1 of the appendix \citep{appendix}):
\begin{align}
    & \CELBO(q(B,\mu, f, \bm u ),p(\mD |B,\mu,  f, \bm u),p(B, \mu, f, \bm u )) \nn \\
&=\underbrace{\E_{q(B,\mu, f)} \Big [ \log p(\mD, B |  f,\mu )\Big ]}_{\text{Data Dependent Expectation (DDE)}}+H_B\\
&\quad -\KL(q(\mu ) || p(\mu ))-\KL(q(\bm u) || p(\bm u)).\label{eq:vbhp_elbo}
\end{align}
where $H_B = -\sum_{i=1}^{N}\sum_{j=0}^{i-1}q_{ij}\log q_{ij}$ is the entropy of the variational posterior $B$. The KL terms are between gamma and Gaussian distributions for which closed forms are provided in A.2 of the appendix \citep{appendix}.

\subsection{Data Dependent Expectation}
Now, we are left with the problem of computing the data dependent expectation (DDE) in Eqn.\eqref{eq:vbhp_elbo}.
The DDE is w.r.t. the variational posterior probability $q(B,\mu,f)$. From Eqn.\eqref{eq:vbhp_approx_joint_posterior}, $q(B,\mu,f) = \int q(B,\mu,f,\bu)\intd \bu = q(B)q(\mu)q(f)$ and $q(f)$ is identical to Eqn.\eqref{eq:post_f}. As a result, we can compute the DDE w.r.t. $q(B)$ first, and then w.r.t. $q(\mu)$ and $q(f)$. 

\textbf{Expectation w.r.t. $q(B)$.}
From Eqn.\eqref{eq:log_db}, we easily obtain $\log p(\mD, B | f, \mu)$ by replacing $\phi$ with $f^2$, whereupon it is clear that only $b_{ij}$ in $\log p(\mD, B | f, \mu)$ is dependent on $B$. Therefore, $\E_{q(B)}[\log p(\mD, B | f, \mu)]$ is computed as:
{\medmuskip=1mu
	\thinmuskip=1mu
	\thickmuskip=1mu
\begin{align}
    &\E_{q(B)}[\log p(\mD, B | f, \mu)]
    \\
    &=\sum_{i=1}^{N}\Big(\sum_{j=1}^{i-1} q_{ij} \log f^2_{ij}+q_{i0}\log \mu \Big) -\sum_{i=1}^{N}\int_{\mT_i}f^2 -\mu |\mT|
\end{align}}
where $f_{ij} \equiv f(\bm x_i - \bm x_j)$.

\textbf{Expectation w.r.t. $q(f)$ and $q(\mu)$.} We compute the expectation w.r.t. $q(f)$ and $q(\mu)$ by exploiting the expectation and the log expectation of the Gamma distribution:  $\E_{q(\mu)}(\mu) = kc$ and $\E_{q(\mu)}(\log (\mu)) = \psi(k)+\log c$, and also  the property $\E(x^2) = \E(x)^2+\Var(x)$:
{\medmuskip=1mu
	\thinmuskip=1mu
	\thickmuskip=1mu
\begin{align}
\text{DDE}&=\sum_{i=1}^{N}\Big [\sum_{j=1}^{i-1} q_{ij} \E_{q(f)}(\log f_{ij}^2)+q_{i0}\big(\psi(k)+\log c\big)
\\
&\quad-\int_{\mT_i} \E_{q(f)}^2(f) -\int_{\mT_i}\Var_{q(f)}(f) \Big ] - kc |\mT|,\label{eq:elbo_DDE}
\end{align}}%
where $\psi$ is the Digamma function. We provide closed form expressions for $\int_{\mT_i} \E_{q(f)}^2(f)$ and $\int_{\mT_i} \Var_{q(f)}(f)$ in A.3 of the appendix \citep{appendix}. As in VBPP, $\E_{q(f)}(\log f_{ij}^2) = -\widetilde{G}(- \nu_{ij}^2/(2 \Sigma_{ij}))+\log (\Sigma_{ij}/2)-C$ is available in the closed form with a hyper-geometric function,
where $\nu_{ij}=\nu(\bm x_i - \bm x_j)$, $ \Sigma_{ij} =  \Sigma(\bm x_i - \bm x_j,\bm x_i - \bm x_j)$ ($\nu$ and $\Sigma$ are defined in Eqn.\eqref{eq:post_f}), $C \approx 0.57721566$ is the Euler-Mascheroni constant and $\widetilde{G}$ is defined as $\widetilde{G}(z) = \prescript{}{1}{F} _1^{(1,0,0)}(0,1/2,z)$, i.e., the partial derivative of the confluent hyper-geometric function $\prescript{}{1}{F} _1^{}$ w.r.t. the first argument. We compute $\widetilde{G}$  using the method of \citet{ancarani2008derivatives} and  implement $\tilde{G}$ and $\tilde{G}'$ by linear interpolation of a lookup table (see a demo in the supplementary material).

\subsection{Predictive Distribution of $\phi$} The predictive distribution of $f(\bm x)$ depends on the posterior $\bu$. We assume that the optimal variational distribution of $\bu$ approximates the true posterior distribution, namely $q(\bu | \mD, {\bm\theta'}^*) = \mathcal N(\bu | \bm m^*, \bS^*)\approx p(\bu | \mD, \bm \theta)$.
Therefore, there is $q(f | \mD, {\bm \theta'}^*)\approx p(f | \mD, \bm \theta)$, i.e., the approximate predictive $f(\tilde{\bm x}) \sim \mathcal N(\bm K_{\tilde{x}z}\bm K_{zz}^{-1}\bm m^{*}, \bm K_{\tilde{x}\tilde{x}}-\bm K_{\tilde{x}z}\bm K_{zz'}^{-1}\bm K_{z\tilde{x}}+\bm K_{\tilde{x}z}\bm K_{zz'}^{-1}\bS^{*} \bm K_{zz'}^{-1} \bm K_{z\tilde{x}})\equiv \mathcal N(\tilde{\nu},\tilde{\sigma}^{2})$. Given the relation $\phi=f^2$, it is straightforward to derive the corresponding $\phi(\tilde{\bm x})\sim \text{Gamma}(\tilde{k},\tilde{c})$ where the shape $\tilde{k}=(\tilde{\nu}^{2}+\tilde{\sigma}^{2})^2/[2\tilde{\sigma}^{2}(2\tilde{\nu}^{2}+\tilde{\sigma}^{2})]$ and the scale $\tilde{c}=2\tilde{\sigma}^{2}(2\tilde{\nu}^{2}+\tilde{\sigma}^{2})/(\tilde{\nu}^{2}+\tilde{\sigma}^{2})$.
%


\section{New Variational Inference Schema}
\label{sec:new_vi}
We now propose a new variational inference (VI) schema which uses a tighter ELBO than the common one, \textit{i.e.}
\begin{theorem}
For VBHP, there is a tighter ELBO
\begin{equation}
\label{eq:tighter_elbo}
   \underbrace{\E_{q(B,\mu, f)} \Big [ \log p(\mD, B |  f,\mu )\Big ]+H_B }_{\equiv \normalfont \text{TELBO}}\leq \log p(\mD).
\end{equation}
\end{theorem}
\begin{remark}
TELBO is tighter because it is equivalent to the CELBO (Eqn.\eqref{eq:vbhp_elbo}) except without subtracting non-negative KL divergences over $\mu$ and $\bu$. Other graphical models such as the variational Gaussian mixture model \citep{attias1999inferring} have a similar TELBO. Later on, we propose a new VI schema based on the TELBO.
\end{remark}
\begin{proof} With the variational posterior probability of the branching structure $q(B)$ defined in Eqn.\eqref{eq:q_b} and through the Jensen's inequality, we have:
\begin{align}
    \log p(\mathcal D)
	&\geq \sum_B q(B) \log p(\mathcal D, B )+ H_B,
	\label{eq:tighter_elbo_jensen}
\end{align}
where $H_B$ is the entropy of $B$ defined in Eqn.\eqref{eq:vbhp_elbo}. The term $\sum_B q(B) \log p(\mathcal D, B )$ can be understood as follows. Consider that infinite branching structures are drawn from $q(B)$ independently, say $\{B_i\}_{i=1}^{\infty}$. Given a branching structure $B_i$, the Hawkes process can be decomposed into a cluster of Poisson processes, denoted as $(\mD, B_i)$, and the corresponding log-likelihood is $\log p(\mD, B_i)$. Then, $\sum_B q(B) \log p(\mathcal D, B )$ is the mean of all log likelihoods $\{\log p(\mD,B_i) \}_{i=1}^{\infty}$,
\begin{align}
\lim_{n \rightarrow \infty} \dfrac{1}{n}\sum_{i=1}^n \log p(\mD,B_i) &= \lim_{n \rightarrow \infty} \sum_{B} \dfrac{n_B}{n}\log p(\mD,B)\\
& = \sum_B q(B) \log p(\mathcal D, B ),\label{eq:tighter_elbo_equality}
\end{align}
where $n_B$ is the number of occurences of branching structure $B$. Since all branching structures $\{B_i\}_{i=1}^{\infty}$ are i.i.d., the clusters of Poisson processes  generated over $\{B_i\}_{i=1}^{\infty}$ should also be independent, i.e., $\{(\mD,B_i)\}_{i=1}^{\infty}$ are i.i.d.. It follows that
\begin{align}
\sum_{i=1}^{\infty}\log p(\mD,B_i)=\log p(\{(\mD,B_i)\}_{i=1}^{\infty}) \label{eq:key_step}.
\end{align}
We compute the CELBO of $\log p(\{(\mD,B_i)\}_{i=1}^{\infty})$ by making $\bm z = (\mu, f, \bu)$ and $\bm x = \{(\mD, B_i)\}_{i=1}^{n}$ in Eqn.\eqref{eq:vi_elbo}:
{\medmuskip=0mu
	\thinmuskip=0mu
	\thickmuskip=0mu
\begin{align}
    \log p(\{(\mD,B_i)\}_{i=1}^{n}) &\geq E_{q(f,\mu)} \big [\log p(\{(\mD, B_i)\}_{i=1}^{n} | f,\mu)]\\
    -\KL&(q(\mu ) || p(\mu ))-\KL(q(\bm u) || p(\bm u)).
    \label{eq:tighter_elbo_elbo}
\end{align}}
Further, we plug Eqn.\eqref{eq:key_step} and Eqn.\eqref{eq:tighter_elbo_elbo} into Eqn.\eqref{eq:tighter_elbo_equality}:
\begin{align}
\text{Eqn.\eqref{eq:tighter_elbo_equality}} &=\lim_{n \rightarrow \infty}\dfrac{1}{n}\log p(\{(\mD,B_i)\}_{i=1}^{n})
\\
&\overset{(a)}{\geq} \lim_{n \rightarrow \infty} \dfrac{1}{n}E_{q(f,\mu)} \big [\log p(\{(\mD, B_i)\}_{i=1}^{n} | f,\mu)] \\
&\quad \overset{(b)}{=} \lim_{n \rightarrow \infty} \sum_{n_B}\dfrac{n_B}{n}E_{q(f,\mu)} \big [\log p(\mD, B| f,\mu)\big ]\\
&\quad = E_{q(f,\mu,B)} \big [\log p(\mD, B| f,\mu)\big ]
\end{align}
where (a) is because the finite values of KL terms are divided by infinitely large $n$, and (b) is due to  i.i.d. $(\mD, B_i)$ and the variational posterior $B$ being independent of $f$ and $\mu$.
Finally, we plug the above inequality into Eqn.\eqref{eq:tighter_elbo_jensen} and obtain the TELBO.
\end{proof}

\paragraph{New Optimization Schema for VBHP} To optimize the model parameters, we employ the expectation-maximization algorithm. Specifically,  in the \textbf{E step}, all $q_{ij}$ are optimized to maximize the CELBO, and in the \textbf{M step}, $\bm m$, $\bm S$, $k$ and $c$ are updated to increase the CELBO. We don't use the TELBO to optimize the variational distributions because it doesn't guarantee minimizing the KL divergence between variational and true posterior distributions.
Instead, the TELBO is employed to select GP hyper-parameters:
\begin{align}
\{\alpha_i^*\}_{i=1}^{R}, \gamma^* = \text{argmax}_{\{\alpha_i\}_{i=1}^{R}, \gamma}  \text{TELBO}.
\end{align}
The TELBO bounds the marginal likelihood more tightly than CELBO, and is therefore expected to lead to a better predictive performance --- an intuition which we empirically validate in Sec.\ref{sec:experiments}.

The updating equations for $q_{ij}$ are derived through maximization of Eqn.\eqref{eq:vbhp_elbo} under the constraints $\sum_{j=0}^{i-1} q_{ij}=1$ for all $i$. This maximization problem is dealt with the Lagrange multiplier method, and yields the below updating equations:
\begin{align}
    q_{ij} &= \begin{cases}
    \exp(\E_{q(f)}(\log f_{ij}^2))/A_i,\quad j>0;
    \\
    \theta\exp(\psi(k))/A_i, \quad j=0,
    \end{cases}
\end{align}
where $A_i=\theta \exp(\psi(k)) + \sum_{j=1}^{i-1}\exp(\E_{q(f )}(\log f_{ij}^2) )$ is the normalizer.

Furthermore, and similarly to VBPP, we fix the inducing points on a regular grid over $\mT$
. Despite the observation that more inducing points lead to better fitting accuracy \citep{lloyd2015variational,snelson2006sparse}, in the case of our more complex VBHP, more inducing points may cause slow convergence (Fig.5a \citep{appendix}) for some hyper-parameters, and therefore lead to poor performance in limited iterations. Generally, more inducing points improve accuracy at the expense of longer fitting time. 


\section{Acceleration Trick} 
\label{sec:acceleration}
\paragraph{Time Complexity Without Acceleration}
In the E step of model optimization, updating $q_{ij}$ requires computing the mean and the variance of all $f_{ij}$, which both take $O(M^3+ M^2N^2)$ with $N$ points in the HP and $M$ inducing points. Here, we omit the dimension of data $R$ since normally $M>R$ for a regular grid of inducing points. Similarly, in the M step, computing the hyper-geometric term requires the means and variances of all the $f_{ij}$. Finally, computation of the integral terms takes $O(M^3N)$. Thus, the total time complexity per iteration is $O(M^3N+M^2N^2)$. 

\paragraph{Acceleration to Linear Time Complexity}
To accelerate our VBHP, similarly to \citet{zhang2018efficient} we exploit the stationarity of the triggering kernel, assuming the kernel has negligible values for sufficiently large inputs. As a result, sufficiently distant pairs of points
do not enter into the computations.
This trick reduces possible parents of a point from all prior points to a set of neighbors. The number of relevant neighbors is bounded by a constant $C$ and as a result\textbf{} the total time complexity is reduced to $O(CM^3N)$.

Specifically, we introduce a compact region $\mathcal S = \bigtimes_{r=1}^{R} [\mathcal
S_{r}^{\min}, \mathcal 
S_{r}^{\max}] \subseteq \mT$ so that $\phi(\bm x_i - \bm x_j)=0$ and $q_{ij}=0$ if $\bm x_i - \bm x_j \not\in \mathcal S$. As a result, all terms related to $\bm x_i - \bm x_j \not\in \mathcal S$ vanish. To choose a suitable $\mathcal S$, we again use the TELBO, taking the smallest $\bm S$ for which the TELBO doesn't drop significantly; we optimize $S_{r}^{\min}$ and $S_{r}^{\max}$ by grid search with other dimensions fixed (so that this step is run $R$ times in total) and we optimize $S_{r}^{\min}$ after optimizing $S_{r}^{\max}$.

Rather than selecting pairs of points in each iteration in the manner of Halpin's trick \citep{halpin2012algorithm,zhang2018efficient}, our method pre-computes those pairs, leading to gains in computational efficiency. The similar aspect is that both tricks have hyper-parameters to select to threshold the triggering kernel value. We employ the TELBO for hyper-parameter selection while frequentist methods use the cross validation.

\section{Experiments}
\label{sec:experiments}
\begin{figure*}[t]
\centering
\subfloat[Variational Posterior Triggering Kernel]{\pdfimageresolution=300\includegraphics[width=0.3\textwidth]{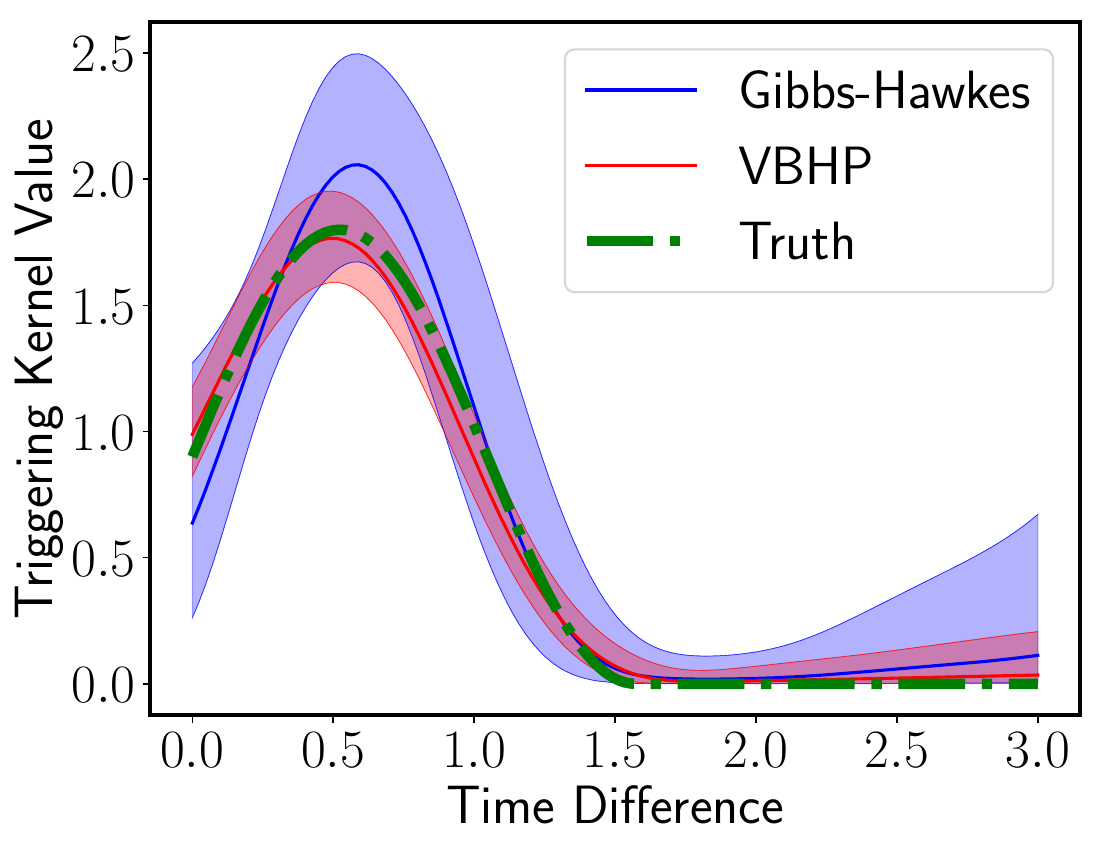} \label{fig:posterior_toy_sin}} 
\subfloat[TELBO]{\pdfimageresolution=300\includegraphics[width=0.32\textwidth]{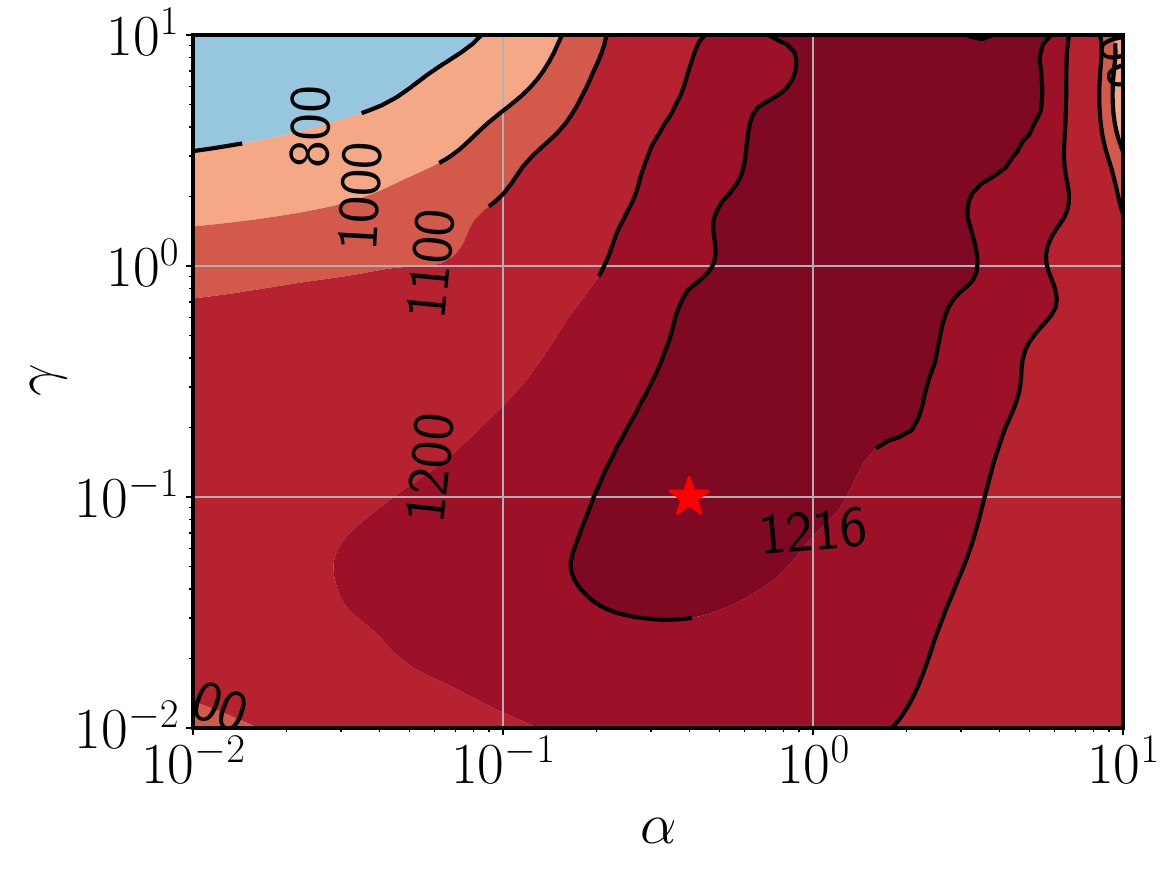}\label{fig:marginal_likelihood_vbhp}}
\subfloat[L2 Distance for $\phi$]{\pdfimageresolution=300\includegraphics[width=0.32\textwidth]{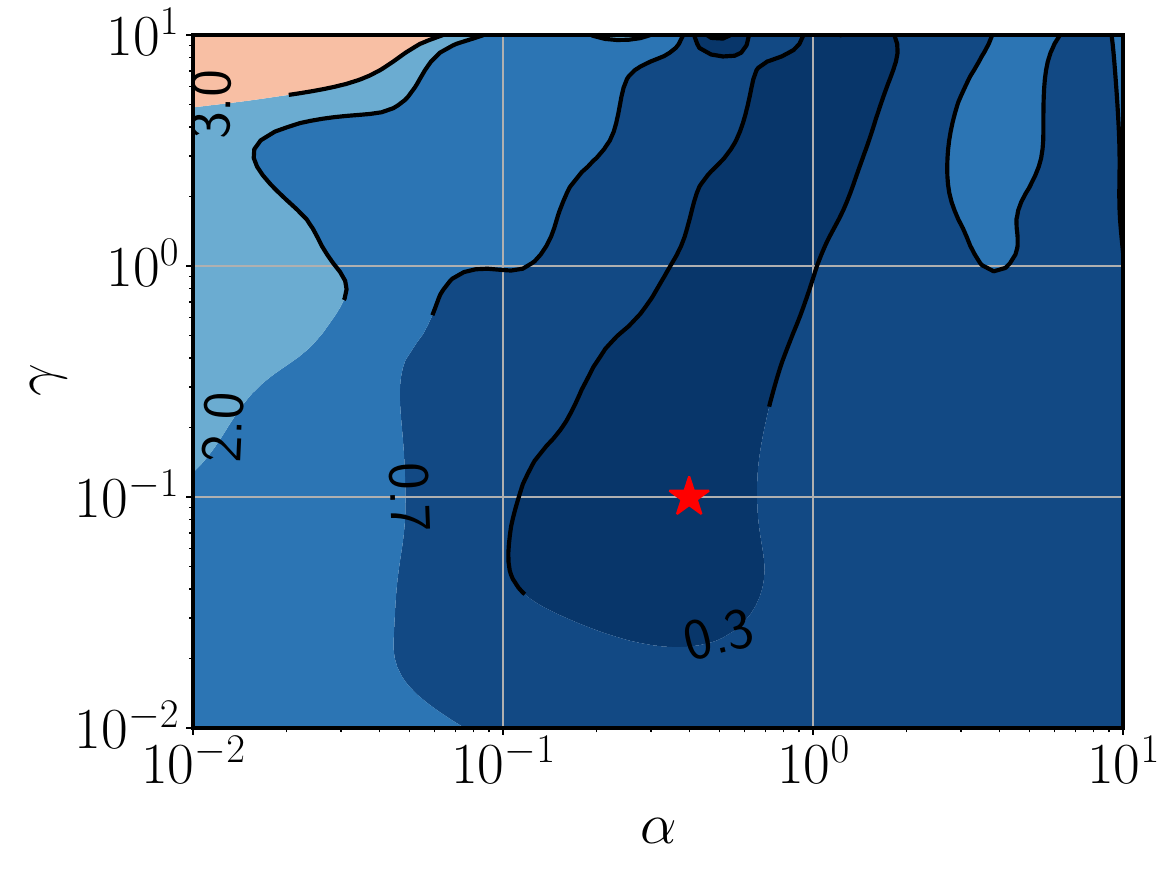}\label{fig:l2f_vbhp}}
\caption{The Relationship between the Log Marginal Likelihood and the $\text{L}_2$ Distance. In (a), the true $\phi_{\sin}$ (dash green) is plotted with the median (solid) and the [0.1, 0.9] interval (filled) of the approximate posterior triggering kernel obtained by VBHP and Gibbs Hawkes (10 inducing points). It uses the maximum point of the TELBO (red star in (b)). In (c), the maximum point of the TELBO is marked. The maximum point overlaps with that of the CELBO. $[0,1.4]$ is used as the support of the predictive triggering kernel and 10 inducing points are used.}
\label{fig:relationship_ml_eva}
\end{figure*}

\begin{figure*}
\centering
\subfloat[Expected TELBO (Train)]{\pdfimageresolution=300\includegraphics[width=0.25\textwidth]{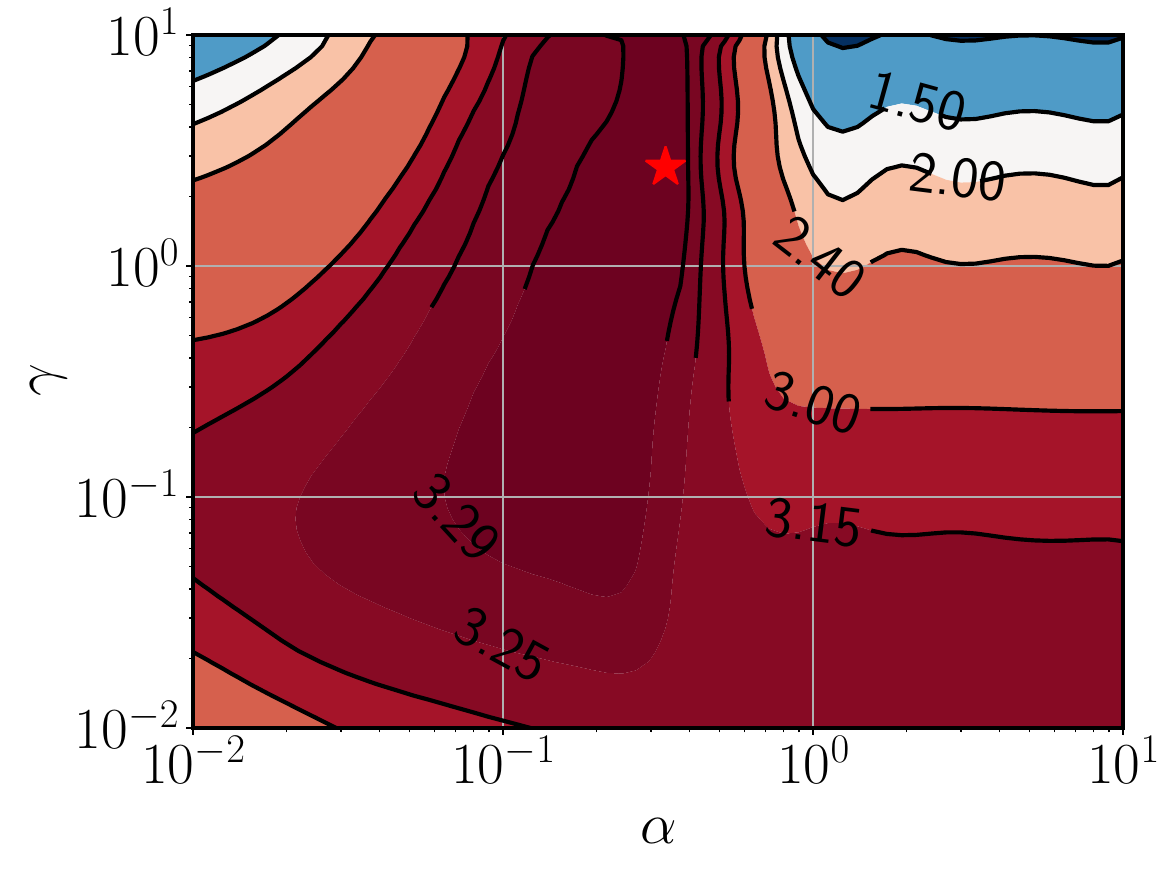}\label{fig:vbhp_telbo_train}}
\subfloat[HLL]{\pdfimageresolution=300\includegraphics[width=0.25\textwidth]{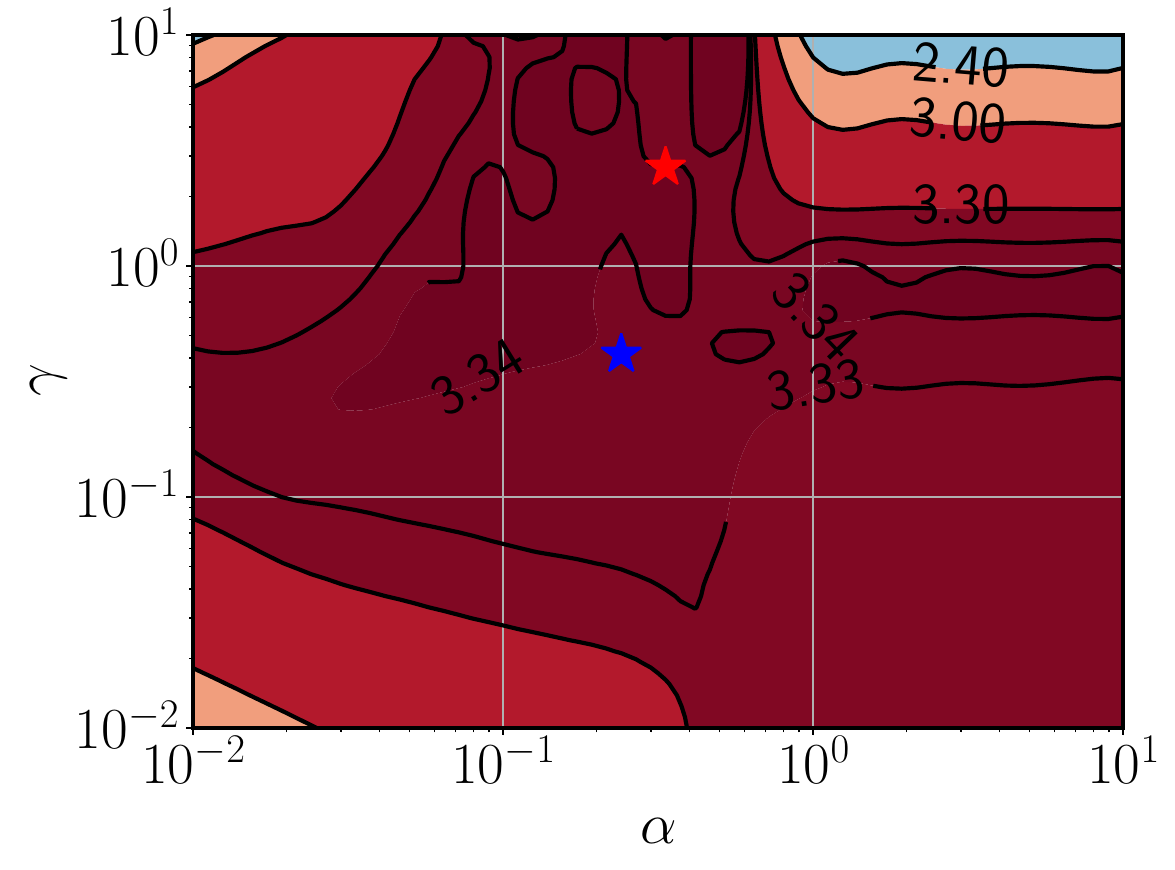}
\label{fig:vbhp_hll_test}}
\subfloat[Time VS Inducing Points]{\pdfimageresolution=300\includegraphics[width=0.24\textwidth]{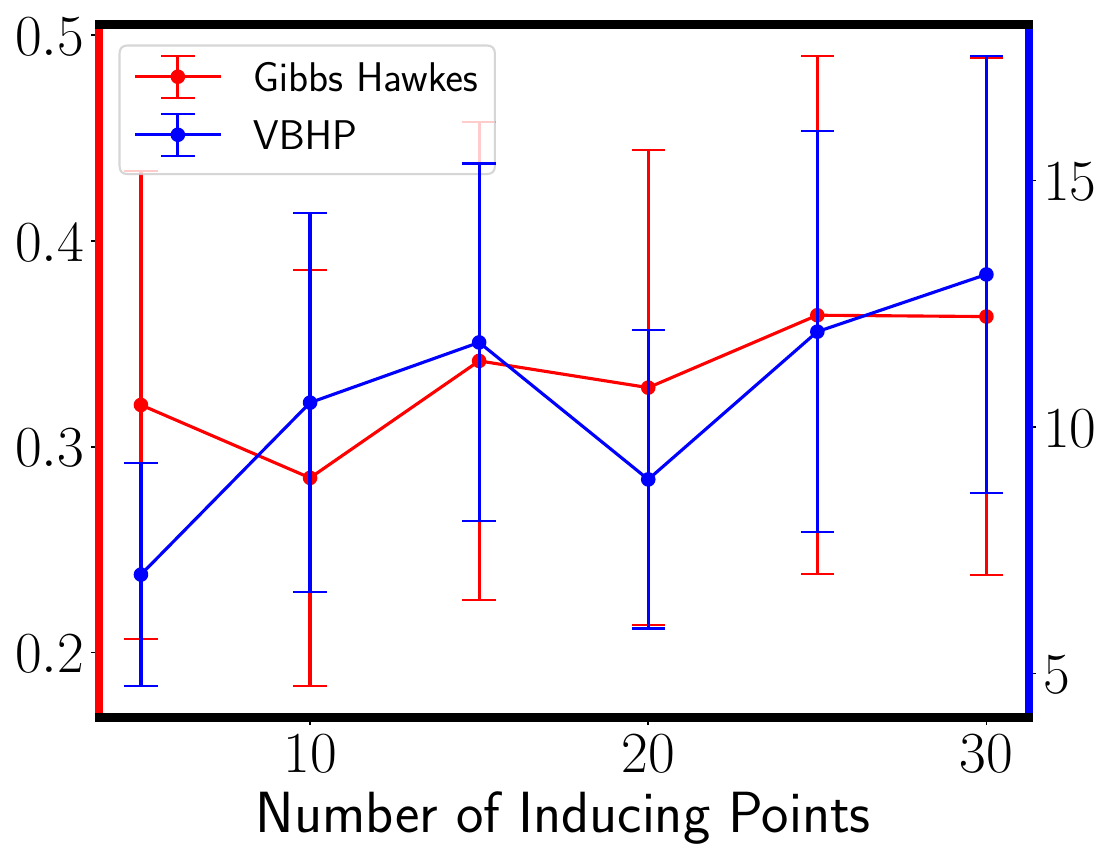} \label{fig:time_vs_inducing_points}}
\subfloat[Time VS Data Size]{\includegraphics[width=0.235\textwidth]{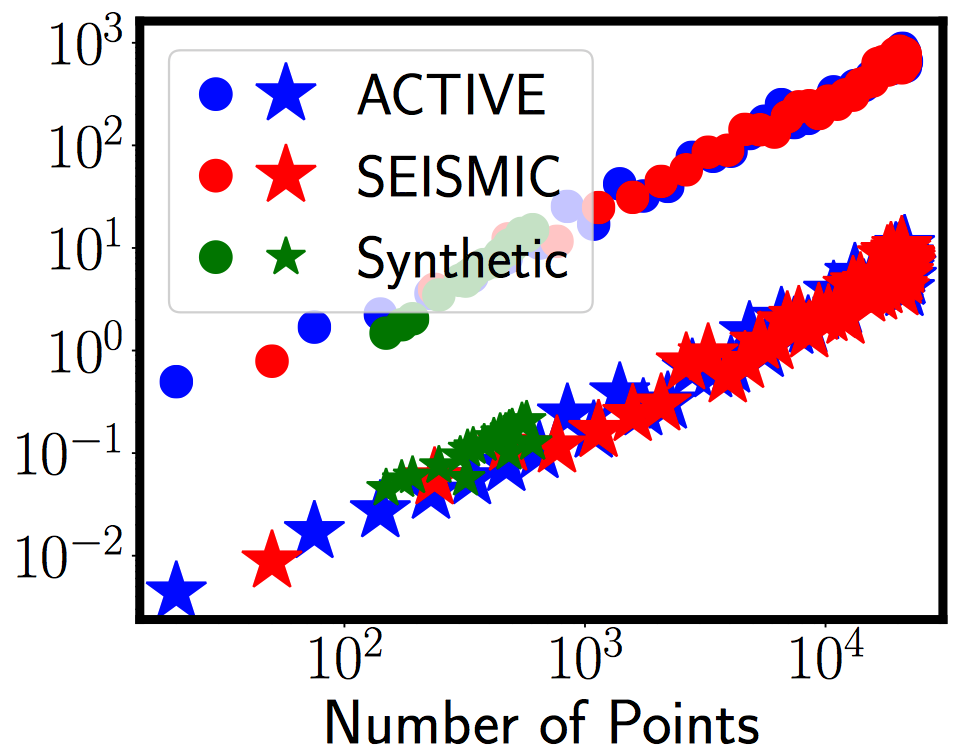} \label{fig:time_vs_data_size}}
\caption{(a)$\sim$(b) The Relationship between the TELBO and the HLL; (c)$\sim$(d) Average Fitting Time (Seconds) Per Iteration. In (a), the maximum point is marked by the red star. In (b), the maximum points of the TELBO and CELBO are marked by red and blue stars. (c) is plotted on 50 processes. (d) shows the fitting time of Gibbs Hawkes (star) and VBHP (circle) on 120 processes. 10 inducing points are used unless specified.}
\end{figure*}

\textbf{Evaluation.} We employ two metrics: the first is the $\mathbf{L_2}$ distance (for cases with a known ground truth), which measures the difference between predictive and truth Hawkes kernels, formulated as $\text{L}_2(\phi_{\text{pred}},\phi_{\text{true}}) = (\int_{\mT} (\phi_{\text{pred}}(\bm x)-\phi_{\text{true}}(\bm x))^2 \intd \bm x)^{0.5}$ and $\text{L}_2(\mu_{\text{pred}},\mu_{\text{true}}) = |\mu_{\text{pred}}-\mu_{\text{true}} |$; the second is the hold-out log likelihood (\textbf{HLL}), which describes how well the predictive model fits the test data, formulated as $\log p(\mD_{\text{Test}}=\{\bm x_i\}_{i=1}^N | \mu, f)= \sum_{i=1}^{N}\log \lambda(\bm x_i)-\int_{\mT} \lambda$. To calculate the HLL for each process, we generate a number of test sequences by every time randomly assigning each  point of the original process to either a training or testing sequence with equal probability; HLLs of test sequences are normalized (by dividing test sequence length) and averaged.

\textbf{Prediction.}
We use the pointwise mode of the approximate posterior triggering kernel as the prediction because it is computationally intractable to find the posterior mode at multiple point locations \citep{zhang2018efficient}. Besides, we  exploit the mode of the approximate posterior background intensity as the predictive background intensity.

\begin{table*}
\centering
\begin{tabular}{cccccccc}
\toprule
Measure & Data & SumExp & ODE & WH & Gibbs Hawkes & VBHP (C) & VBHP (T)
\\ \midrule
\multirow{6}{*}{$\text{L}_2$}    &\multirow{2}{*}{Sin} &   $\phi$:$0.693_{\pm0.028}$ &  $0.665_{\pm0.121}$   & $2.463_{\pm0.145}$   & $0.408_{\pm0.198}$ & $\bm{0.152}_{\pm0.091}$ &  $0.183_{\pm0.076}$
\\ 
 & & $\mu$:$2.968_{\pm1.640}$ & $4.514_{\pm3.808}$   & $6.794_{\pm5.054}$   &  $4.108_{\pm3.949}$ & $0.640_{\pm0.528}$ &$\bm{0.579}_{\pm0.523}$
\\ 
 & \multirow{2}{*}{Cos} &  $\phi$:$0.473_{\pm0.102}$       & $0.697_{\pm0.065}$  & $1.743_{\pm0.083}$  & $0.667_{\pm0.686}$ &  $0.325_{\pm0.073}$ &  $\bm{0.292}_{\pm0.096}$
\\ 
&    &  $\mu$:$2.751_{\pm1.902}$ &  $7.030_{\pm5.662}$   & $6.099_{\pm4.613}$& $4.685_{\pm4.421}$      &       $0.555_{\pm0.294}$ & $\bm{0.515}_{\pm0.293}$
\\ 
& \multirow{2}{*}{Exp} &  $\phi$:$\bm{0.133}_{\pm0.138}$ & $1.835_{\pm0.539}$   & $2.254_{\pm2.042}$   &    $0.676_{\pm0.233}$ &$0.257_{\pm0.086}$ &$0.235_{\pm0.102}$
\\
&    & $\mu$:$3.290_{\pm1.991}$   & $8.969_{\pm8.604}$   & $16.66_{\pm20.95}$& $7.648_{\pm9.647}$& $\bm{0.471}_{\pm0.432}$& $0.486_{\pm0.418}$    
\\ \midrule
\multirow{5}{*}{HLL} & Sin   &  $3.490_{\pm0.400}$ & $3.489_{\pm0.413}$ &  $3.233_{\pm0.273}$  &      $3.492_{\pm0.406}$    & $3.488_{\pm0.400}$    & $\bm{3.497}_{\pm0.406}$
\\
& Cos &  $3.874_{\pm0.544}$ & $3.872_{\pm0.552}$     &$3.613_{\pm0.373}$ & $3.871_{\pm0.562}$ & $3.876_{\pm0.541}$ & $\bm{3.878}_{\pm0.548}$
\\ 
& Exp &  $2.825_{\pm0.481}$ &   $2.822_{\pm0.496}$ &    $2.782_{\pm0.490}$  &  $2.826_{\pm0.492}$       & $2.826_{\pm0.491}$    & $\bm{2.829}_{\pm0.487}$
\\ 
& ACTIVE  &   $1.692_{\pm1.371}$ & $0.880_{\pm2.716}$    &$0.710_{\pm0.943}$  & $1.323_{\pm2.160}$      &   $1.824_{\pm1.159}$ &   $\bm{1.867}_{\pm1.181}$
\\
& SEISMIC & $2.943_{\pm0.959}$ & $2.582_{\pm1.665}$ & $1.489_{\pm1.796}$ & $3.110_{\pm1.251}$  & $3.143_{\pm0.895}$ & $\bm{3.164}_{\pm0.843}$  \\
\bottomrule
\end{tabular}
\caption{Results on Synthetic and Real World Data (Mean $\pm$ One Standard Variance). VBHP (C) and (T) use the CELBO and the TELBO to update the hyper-parameters respectively.}
\label{tab:results}
\end{table*}

\textbf{Baselines.}
We use the following models as baselines. \textbf{(1)} A parametric Hawkes process  equipped with the sum of exponential (\textbf{SumExp}) triggering kernel $\phi(\bm x) =\sum_{i=1}^{K} a_1^i a_2^i \exp(-a_2^i \bm x)$ and the constant background intensity. \textbf{(2)} The ordinary differential equation (\textbf{ODE}) based non-parametric non-Bayesian Hawkes process \citep{zhou2013learning}. The code is publicly available \citep{2017arXiv170703003B}.
\textbf{(3)} Wiener-Hopf (\textbf{WH}) equation based non-parametric non-Bayesian Hawkes process \citep{zhou2013learning}. The
code is publicly available \citep{2017arXiv170703003B}.
\textbf{(4)} The Gibbs sampling based Bayesian non-parametric Hawkes process (\textbf{Gibbs Hawkes}) \citep{zhang2018efficient}. For fairness, the ARD kernel is used by Gibbs Hawkes and corresponding eigenfunctions are approximated by Nystr\"om method \citep{NIPS2000_1866}, where regular grid points are used as VBHP. Different from batch training in \citep{zhang2018efficient}, all experiments are conduct on single sequences.

\subsection{Synthetic Experiments}
\label{sec:synthetic_data}
\textbf{Synthetic Data.} Our synthetic data are generated from three Hawkes processes over $\mT = [0, \pi]$, whose triggering kernels are sin, cos and exp functions respectively, shown as below, and whose background intensities are the same $\mu=10$:
\begin{align}
&\phi_{\sin}(x)= 
0.9[\sin(3x)+1], x\in [0,\pi/2]; \text{otherwise}, 0; \label{eq:sin_tk} \\
&\phi_{\cos}(x)=\cos(2x)+1,x\in [0,\pi/2]; \text{otherwise}, 0; \label{eq:cos_tk} \\
&\phi_{\exp}(x)=5\exp(-5x), x\in [0,\infty).\label{eq:exp_tk}
\end{align}
As a result, for any generated sequence, say $\{x_i\}_{i=1}^{N}$, $\mT_i = [0, \pi - x_i]$ is used in the CELBO and the TELBO.

\textbf{Model Selection.}
As the marginal likelihood $p(\mD | \bm \theta)$ is a key advantage of our method over non-Bayesian approaches \citep{zhou2013learning,bacry2014second}, we investigate  its efficacy for model selection. Fig.\ref{fig:marginal_likelihood_vbhp} shows the contour plot of the approximate log marginal likelihood (the TELBO) of a sequence. It is observed that the contour plot of the TELBO has agreement to the contour plots of $\text{L}_2(\phi)$ (Fig.\ref{fig:l2f_vbhp}) --- GP hyper-parameters with relatively high marginal likelihoods have relatively low $\text{L}_2$ errors. Fig.\ref{fig:posterior_toy_sin} plots the posterior triggering kernel corresponding to the maximal approximate marginal likelihood. 
Similar agreement is also observed between the TELBO and the HLL (Fig.\ref{fig:vbhp_telbo_train}, \ref{fig:vbhp_hll_test}). This demonstrates the practical utility of both the marginal likelihood itself and our approximation of it. 

\textbf{Evaluation.} To evaluate VBHP on synthetic data, 20 sequences are drawn from each model and 100 pairs of train and test sequences drawn from each sample to compute the HLL. We select GP hyper-parameters of Gibbs Hawkes and of VBHP by maximizing approximate marginal likelihoods.  Table \ref{tab:results} shows evaluations for baselines and VBHP (using 10 inducing points for trade-off between accuracy and time, so does Gibbs Hawkes) in both $\text{L}_2$ and HLL. VBHP achieves the best performance but is two orders of magnitudes slower than Gibbs Hawkes per iteration (shown as Fig.\ref{fig:time_vs_inducing_points} and \ref{fig:time_vs_data_size}). The TELBO performs closely to the CELBO in the $\text{L}_2$ error and this is also reflected in Fig.\ref{fig:l2f_vbhp} where the maximum points of the TELBO and the CELBO overlap. In contrast, the TELBO consistently improves the performance of VBHP in the HLL, which is also reflected in Fig.\ref{fig:vbhp_hll_test} where hyper-parameters selected by the TELBO tend to have a higher HLL.
Interestingly, when the parametric model SumExp uses the same triggering kernel (a single exponential function) as the ground truth $\phi_{\exp}$, SumExp fits $\phi_{\exp}$ best in $\text{L}_2$ distance while due to learning on single sequences, the background intensity has relatively high errors. Although our method is not aware of the parametric family of the ground truth, it performs well. Compared with non-parametric frequentist methods which have strong fitting capacity but suffer from noisy data and have difficulties with hyper-parameter selection, our Bayesian solution overcomes these disadvantages and achieves better performance.

\subsection{Real World Experiments}
\textbf{Real World Data.} We conclude our experiments with two large scale tweet datasets. \textbf{ACTIVE} \citep{rizoiu2018sir} is a tweet dataset which was collected in 2014 and contains $\sim$41k (re)tweet temporal point processes with links to Youtube videos. Each sequence contains at least 20 (re)tweets. \textbf{SEISMIC} \citep{zhao2015seismic} is a large scale tweet dataset which was collected from October 7 to November 7, 2011, and contains $\sim$166k (re)tweet temporal point processes. Each sequence contains at least 50 (re)tweets.

\textbf{Evaluation.}
Similarly to synthetic experiments, we evaluate the fitting performance by averaging HLL of 20 test sequences randomly drawn from each original datum. We scale all original data to $\mT=[0, \pi]$ (leading to $\mT_i = [0, \pi - x_i]$ used in the CELBO and the TELBO for a sequence $\{x_i\}_{i=1}^{N}$) and employ 10 inducing points to balance time and accuracy. The model selection is performed by maximizing the approximate marginal likelihood. The obtained results are shown in Table \ref{tab:results}. Again, we observe similar predictive performance of VBHP: the TELBO performs better the CELBO; VBHP achieves best scores. This demonstrates our Bayesian model and novel VI schema are useful for flexible real life data.

\textbf{Fitting Time.} We further evaluate the fitting speed\footnote{The CPU we use is Intel(R) Core(TM) i7-7700 CPU @ 3.60GHz and the language is Python 3.6.5.} of VBHP and Gibbs Hawkes on  synthetic and real world point processes, which is summarized in Fig.\ref{fig:time_vs_inducing_points} and 
\ref{fig:time_vs_data_size}. The fitting time is averaged over iterations and we observe that the increasing trends with the number of inducing points and with the data size are similar between Gibbs Hawkes and VBHP. Although VBHP is significantly slower than Gibbs Hawkes per iteration, VBHP converges faster, in 10$\sim$20 iterations (Fig.5 \citep{appendix}), giving an average convergence time of 549 seconds for a sequence of 1000 events, compared to 699 seconds for Gibbs Hawkes. The slope of VBHP in Fig.\ref{fig:time_vs_data_size}  is 1.04 (log-scale) and the correlation coefficient is 0.96, so we conclude that the fitting time is linear to the data size. 

\section{Conclusions}
In this paper, we presented a new Bayesian non-parametric Hawkes process whose triggering kernel is modulated by a sparse Gaussian process and background intensity is Gamma distributed. We provided a novel VI schema for such a model: we employed the branching structure so that the common ELBO is maximized by the expectation-maximization algorithm; we contributed a tighter ELBO which performs better in model selection than the common one. To address the difficulty with scaling with the data size, we utilize the stationarity of the triggering kernel to reduce the number of possible parents for each point. Different from prior acceleration methods, ours enjoys higher efficiency.
On synthetic data and two large Twitter diffusion datasets, VBHP enjoys linear fitting time with the data size and fast convergence rate, and provides more accurate predictions than those of state-of-the-art approaches. The novel ELBO is also demonstrated to exceed the common one in model selection.

\section{Acknowledgements}
We would like to thank Dongwoo Kim for helpful discussions. Rui is supported by the Australian National University and Data61 CSIRO PhD scholarships. 

\bibliography{bibtex}
\bibliographystyle{aaai}
%
\clearpage
\appendix
\onecolumn

Accompanying the submission \textit{Variational Inference for Sparse Gaussian Process Modulated Hawkes Process}.

\section{Omitted Derivations}

\subsection{Deriving Eqn.\eqref{eq:vbhp_elbo}} \label{sec:derive_vbhp_elbo}

As per Eqn.\eqref{eq:vi_elbo}, there is
\begin{align}
    & \CELBO(q(B,\mu, f, \bm u ),p(\mD |B,\mu,  f, \bm u),p(B, \mu, f, \bm u ))   \\
& = \E_{q(B, \mu, f)}  [ \log p(\mD|B, \mu, f) ] - \KL(q(B, \mu, f, \bm u ) || p(B, \mu, f, \bm u ))  
\end{align}
where the KL term can be simplified as
\begin{align} 
    &\KL  ( q(B, \mu, f, \bm u ) || p(B, \mu, f, \bm u)  )   
    \\
    &=\sum_{B} \int\int \int q(B, \mu, f, \bm u) \log \dfrac{q(B )q(\mu)p(f | \bm u)q(\bm u)}{p(B)p(\mu)p(f | \bm u)p(\bm u)}  \intd \bm u \intd f \intd \mu \quad \text{(Eqn.\eqref{eq:vbhp_approx_joint_posterior} and Bayes' rule)}  
    \\
    &=\sum_{B} \int \int \int q(B,\mu, f, \bm u)\intd \bm u \intd f \intd \mu \log \dfrac{q(B)}{p(B) }  
    \\
    &\quad +\sum_{B} \int \int \int
    q(B, \mu, f, \bm u)\intd \bm u \intd f \log \dfrac{q(\mu)}{p(\mu) }  \intd \mu   
    \\
    &\quad  + \sum_{B} \int \int \int q(B, \mu, f, \bm u) \intd f \intd \mu \log \dfrac{q(\bm u)}{p(\bm u)}  \intd \bm u \quad \text{(simplification)}   \\
    & = \sum_{B}q(B) \log \dfrac{q(B)}{p(B) }+\int q(\mu) \log \dfrac{q(\mu)}{p(\mu)}  \intd \mu+\int q(\bm u) \log \dfrac{q(\bm u)}{p(\bm u)}  \intd \bm u \quad \text{(simplification)}   
    \\
    & = \KL(q(B) || p(B ))+\KL(q(\mu ) || p(\mu ))+\KL(q(\bm u ) || p(\bm u ))   .
\end{align}
We utilise the likelihood $p(\mD, B| \mu, f)$ by the reconstruction term and the KL term w.r.t. $B$
\begin{align}
    &\E_{q(B,\mu, f)}  [ \log p(\mathcal D|B,\mu, f) ]-\KL(q(B ) || p(B ))   
    \\
    &= \sum_{B} \int \int q(B,\mu, f) \log p(\mathcal D|B,\mu, f) \intd f \intd \mu -\sum_{B}  q(B ) \log \dfrac{q(B )}{p(B)} \quad \text{(definition)}  
    \\
    &= \sum_{B} \int \int q(B,\mu, f) \log p(\mathcal D|B,\mu, f) \intd f \intd \mu-\sum_{B}  \int \int q(B,\mu, f) \log \dfrac{q(B )}{p(B)} \intd f \intd \mu
    \\
    &\hspace{9cm}\text{(align probabilities)}  
    \\
    &= \sum_{B} \int \int q(B,\mu, f) \log \dfrac{p(\mathcal D|B,\mu, f) p(B)}{q(B )} \intd f \intd \mu \quad \text{(merge)}   
    \\
    &= \sum_{B} \int \int q(B,\mu, f) \log p(\mathcal D,B|\mu, f) \intd f \intd \mu -\sum_B q(B) \log q(B) \quad \text{(merge)}  
    \\
    &= \E_{q(B,\mu, f)} \Big [ \log p(\mathcal D, B| \mu ,f )\Big ] + H_B
\end{align}
where $H_B=-\sum_B q(B) \log q(B)$ is the entropy of $B$ and further computed as follows. We adopt $q(B)$ from Eqn.\eqref{eq:q_b} and the close form expression of $H_B$ is derived as
\begin{align}
H_B 
&=-\sum_{\{\bm b_i\}_{i=1}^{N}} \Big (\prod_{i=1}^{N}\prod_{j=0}^{i-1} q_{ij}^{b_{ij}} \Big )\log \Big (\prod_{i=1}^{N}\prod_{j=0}^{i-1} q_{ij}^{b_{ij}}\Big )
\\
&= -\sum_{j=0}^{k-1}\sum_{\{\bm b_i\}_{i\neq k}} \Big (q_{kj}\prod_{i\neq k}\prod_{j=0}^{i-1}q_{ij}^{b_{ij}} \Big ) \log \Big (q_{kj}\prod_{i\neq k}\prod_{j=0}^{i-1}q_{ij}^{b_{ij}} \Big ) \quad \text{(split the summation)}
\\
&= -\sum_{j=0}^{k-1}\sum_{\{\bm b_i\}_{i\neq k}} \Big (q_{kj}\prod_{i\neq k}\prod_{j=0}^{i-1}q_{ij}^{b_{ij}} \Big ) \Big[\log q_{kj}+\log \Big(\prod_{i\neq k}\prod_{j=0}^{i-1}q_{ij}^{b_{ij}} \Big )\Big] \quad \text{(split the log)}
\\
&=-\sum_{j=0}^{k-1}q_{kj}\log q_{kj}\underbrace{\sum_{\{\bm b_i\}_{i\neq k}} \Big (\prod_{i\neq k}\prod_{j=0}^{i-1}q_{ij}^{b_{ij}} \Big )}_{=1} -\underbrace{\sum_{j=0}^{k-1}q_{kj}}_{=1}\sum_{\{\bm b_i\}_{i\neq k}}\Big (\prod_{i\neq k}\prod_{j=0}^{i-1}q_{ij}^{b_{ij}} \Big )\log \Big(\prod_{i\neq k}\prod_{j=0}^{i-1}q_{ij}^{b_{ij}} \Big ) 
\\
& \hspace{9cm} \text{(distributive law of multiplication)}
\\
&=-\sum_{j=0}^{k-1}q_{kj}\log q_{kj} -\sum_{\{\bm b_i\}_{i\neq k}}\Big (\prod_{i\neq k}\prod_{j=0}^{i-1}q_{ij}^{b_{ij}} \Big )\log \Big(\prod_{i\neq k}\prod_{j=0}^{i-1}q_{ij}^{b_{ij}} \Big )  \quad \text{(simplification)}
\\
&=\cdots \quad \text{(do the same operations as above)}
\\
&=-\sum_{i=1}^{N}\sum_{j=0}^{i-1}q_{ij}\log q_{ij}.
\end{align}

\subsection{Closed Forms of KL Terms in the ELBO}
\label{sec:kl_close_forms}
\begin{align}
 \KL(q(\mu ) || p(\mu)) &= (k-k_0)\psi(k)-k_0\log( c/c_0) -k-\log[\Gamma(k)/\Gamma(k_0)]
+ck/c_0   
\\
\KL(q(\bm u) || p(\bm u)) &
= [\Tr(\bm K_{zz'}^{-1}\bm{S}) + \log|\bm K_{zz'}|/|\bm S|-M+\bm m^T\bm K_{zz'}^{-1} \bm m ]/2
\end{align}

\subsection{Extra Closed Form Expressions for  Eqn.\eqref{eq:elbo_DDE}}

\label{sec:extra_cfe_dde}
\begin{align}
\int_{\mT_i} \E_{q(f)}^2(f)& =\int_{\mT_i} \bm m^T \bm K_{zz'}^{-1} \bm K_{zx} \bm K_{xz} \bm K_{zz'}^{-1} \bm m  \intd  \bm x   
\\
& = \bm m^T \bm K_{zz'}^{-1} \Big (\int_{\mT_i} \bm K_{zx}  \bm K_{xz} \intd  \bm x \Big) \bm K_{zz'}^{-1} \bm m    
\\
& \equiv \bm m^T \bm K_{zz'}^{-1} \bm \Psi_i \bm K_{zz'}^{-1} \bm m.
\end{align}

\begin{align}
\int_{\mT_i} \Var_{q(f)}[f]
& =\int_{\mT_i} \bm K_{xx}-\bm K_{xz}\bm K_{zz'}^{-1}\bm K_{zx}+\bm K_{xz} \bm K_{zz'}^{-1}\bm S \bm K_{zz'}^{-1} \bm K_{zx}  \intd  \bm x   
\\
&=\int_{\mT_i} \gamma -\Tr(\bm K_{zz'}^{-1}\bm K_{zx}\bm K_{xz})+\Tr( \bm K_{zz'}^{-1}\bm S \bm K_{zz'}^{-1} \bm K_{zx}\bm K_{xz})  \intd  \bm x    
\\
&=\int_{\mT_i} \gamma \intd \bm x -\Tr(\bm K_{zz'}^{-1}\int_{\mT_i} \bm K_{zx}\bm K_{xz} \intd \bm x)+\Tr( \bm K_{zz'}^{-1}\bm S \bm K_{zz'}^{-1} \int_{\mT_i} \bm K_{zx}\bm K_{xz} \intd \bm x)    
\\
&\equiv \gamma |\mT_i| -\Tr(\bm K_{zz'}^{-1}\bm \Psi_i)+\Tr(\bm K_{zz'}^{-1}\bm S \bm K_{zz'}^{-1} \bm \Psi_i).
\end{align}

\begin{align}
    \bm \Psi_i(\bm z, \bm z')
    &=\int_{\mT_i} \bm K_{zx}\bm K_{xz'} \intd \bm x 
    \\
    &=\int_{\mT_i} \gamma^2 \prod_{r=1}^{R} \exp \Big( -\dfrac{(x_r - z_r)^2}{2\alpha_r} \Big )\exp \Big( -\dfrac{(z_r'-x_r)^2}{2\alpha_r} \Big ) \intd \bm x  
     \\
    &=\int_{\mT_i} \gamma^2 \prod_{r=1}^{R} \exp \Big( -\dfrac{(z_r - z_r')^2}{4\alpha_r} \Big )\exp \Big( -\dfrac{(\bar z_r-x_r)^2}{\alpha_r} \Big ) \intd \bm x  
     \\
    &=\gamma^2 \prod_{r=1}^{R} \exp \Big( -\dfrac{(z_r - z_r')^2}{4\alpha_r} \Big ) \int_{\mathcal T_{i,r}}  \exp \Big( -\dfrac{(\bar z_r-x_r)^2}{\alpha_r} \Big ) \intd x_r    \quad (y_r = (\bar{z}_r-x_r)/\alpha_r)
     \\
    &=\gamma^2 \prod_{r=1}^{R} -\sqrt{\alpha_r} \exp \Big( -\dfrac{(z_r - z_r')^2}{4\alpha_r} \Big ) \int_{(\bar z_r-\mT_{i,r}^{\min})/\sqrt{\alpha_r}}^{(\bar z_r-\mT_{i,r}^{\max})/\sqrt{\alpha_r}}  \exp ( -y_r^2 ) \intd y_r  
     \\
    &=\gamma^2 \prod_{r=1}^{R} -\dfrac{\sqrt{\pi\alpha_r}}{2} \exp \Big( -\dfrac{(z_r - z_r')^2}{4\alpha_r} \Big )  \Big [\text{erf}\Big (\dfrac{\bar z_r-\mT_{i,r}^{\max}}{\sqrt{\alpha_r}}\Big )-\text{erf}\Big (\dfrac{\bar z_r-\mT_{i,r}^{\min}}{\sqrt{\alpha_r}} \Big )\Big ]  .
\end{align}
 where we explicitly express $\mT_i$ as a Cartesian product $\mT_{i} \equiv \bigtimes_{r=1}^{R} [\mT_{i,r}^{\min}, \mT_{i,r}^{\max}]$ for $R$-dimensional data.
\section{Additional Experiment Results}

\begin{figure*}[h]
    \subfloat[Exp]{\includegraphics[width=0.45\textwidth]{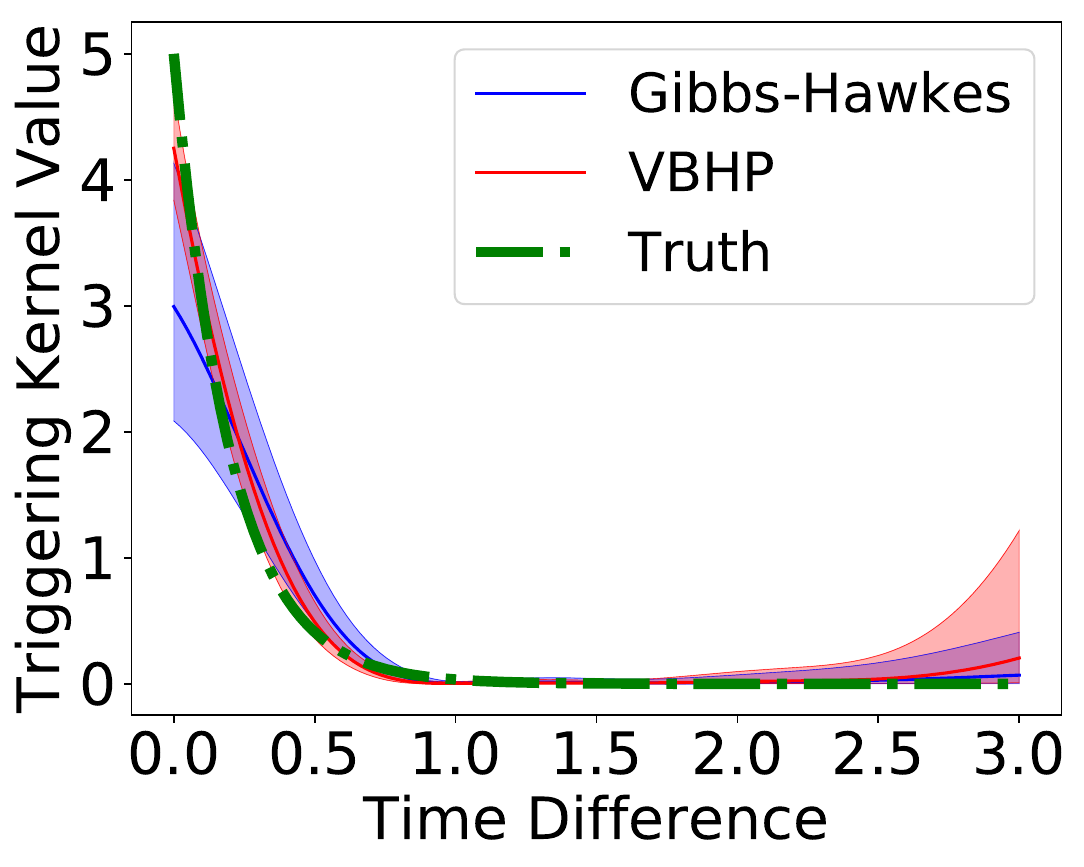}}
    \hfill
    \subfloat[Cos]{\includegraphics[width=0.45\textwidth]{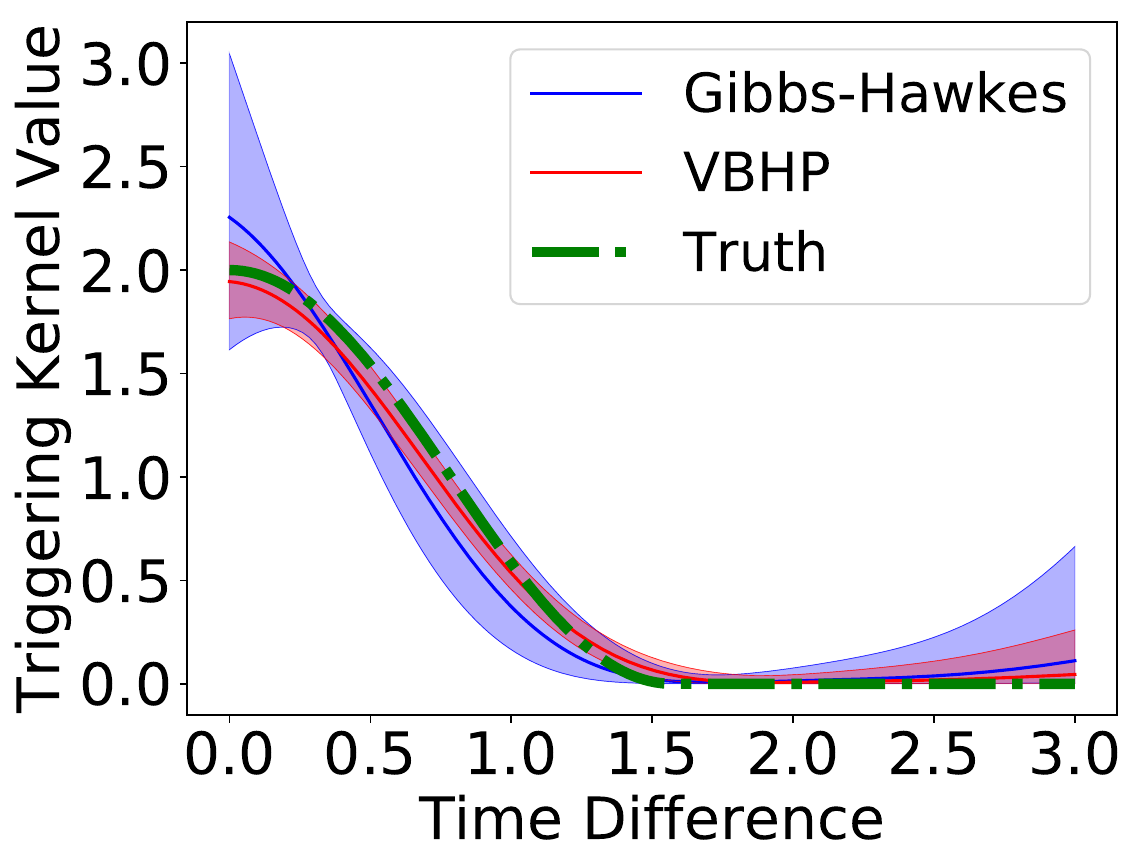}}
    \caption{Posterior Triggering Kernels Inferred By VBHP and Gibbs Hawkes. Results of Gibbs Hawkes are obtained in 2000 iterations.}
\end{figure*}

\begin{figure*}[h]
    \subfloat[VBHP]{\includegraphics[width=0.45\textwidth]{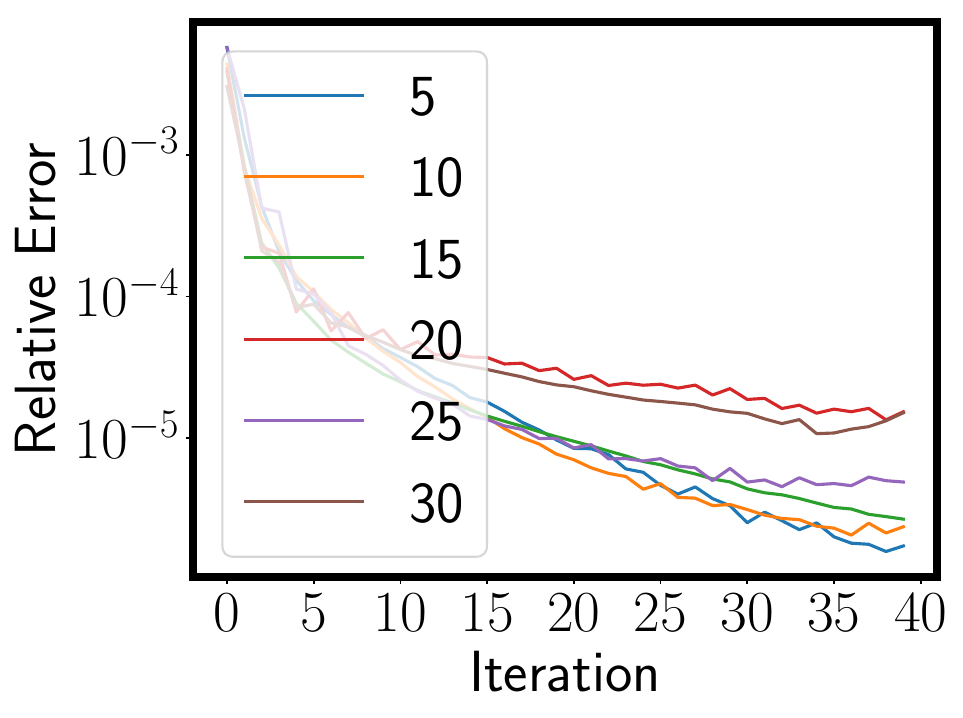}\label{fig:convergence_vbhp}}
    \hfill
    \subfloat[Gibbs Hawkes]{\includegraphics[width=0.45\textwidth]{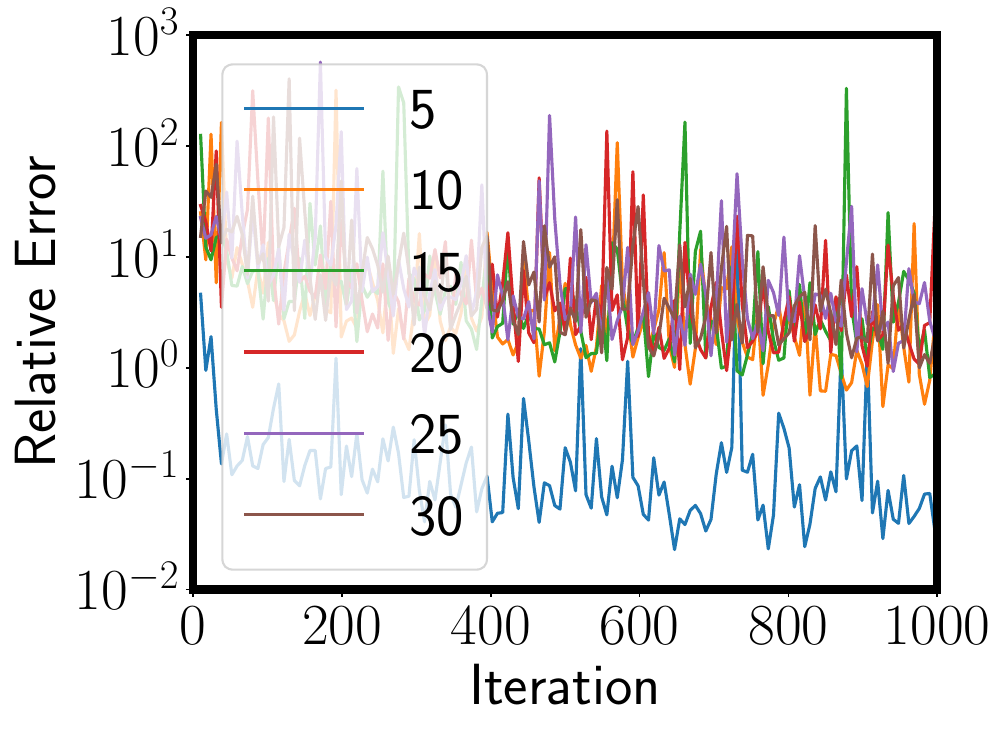}}
    \caption{Convergence Rate of VBHP and Gibbs Hawkes with Different Numbers of Inducing Points. VBHP and Gibbs Hawkes measure respectively the relative error of the approximate marginal likelihood and of the posterior distribution of the Gaussian process.}
    \label{fig:convergence}
\end{figure*}
\end{document}